%% file: main.tex
\pdfoutput=1

\documentclass[11pt]{article}

\usepackage[]{acl}

\usepackage{times}
\usepackage{latexsym}
\usepackage{multirow}
\usepackage{graphicx}
\usepackage{booktabs}
\usepackage{enumitem}
\usepackage{algorithm}
\usepackage{algorithmic}
\usepackage{amsmath}
\usepackage{bm}
\usepackage{colortbl}

\newtheorem{definition}{Definition}

\usepackage[T1]{fontenc}

\usepackage[utf8]{inputenc}

\usepackage{microtype}

\usepackage{inconsolata}

\def\method{DEMO}
\title{\method{}: A Statistical Perspective for Efficient Image-Text Matching}

\makeatletter
\renewcommand{\@fnsymbol}[1]{^\dagger}
\makeatother

\author{Fan Zhang{$^1$},~Xian-Sheng Hua{$^2$},~Chong Chen{$^2$},~Xiao Luo$^3$\thanks{~ Corresponding author.}\\
  {$^1$Georgia Tech Shenzhen Institute, Tianjin University (GTSI)} \\ {$^2$Terminus Group} {$^3$University of California, Los Angeles}\\
  {\tt \small fanzhang@gatech.edu, huaxiansheng@gmail.com, chenchong.cz@gmail.com, xiaoluo@cs.ucla.edu}
}

\begin{document}
\maketitle
\begin{abstract}
\input{1_abstract}
\end{abstract}

\input{2_intro}

\input{3_related}

\input{4_method}

\input{5_experiment}
\input{6_conclusion}


\bibliography{custom}

\input{7_appendix}

\end{document}

%% file: 1_abstract.tex
Image-text matching has been a long-standing problem, which seeks to connect vision and language through semantic understanding. Due to the capability to manage large-scale raw data, unsupervised hashing-based approaches have gained prominence recently. They typically construct a semantic similarity structure using the natural distance, which subsequently provides guidance to the model optimization process. However, the similarity structure could be biased at the boundaries of semantic distributions, causing error accumulation during sequential optimization. To tackle this, we introduce a novel hashing approach termed \underline{D}istribution-based Structur\underline{e} \underline{M}ining with C\underline{o}nsistency Learning (\method{}) for efficient image-text matching. From a statistical view, \method{} characterizes each image using multiple augmented views, which are considered as samples drawn from its intrinsic semantic distribution. Then, we employ a non-parametric distribution divergence to ensure a robust and precise similarity structure. In addition, we introduce collaborative consistency learning which not only preserves the similarity structure in the Hamming space but also encourages consistency between retrieval distribution from different directions in a self-supervised manner. Through extensive experiments on three benchmark image-text matching datasets, we demonstrate that \method{} achieves superior performance compared with many state-of-the-art methods. 

%% file: 2_intro.tex
\section{Introduction}

\begin{figure}[t]
    \centering
    \includegraphics[width=\linewidth]{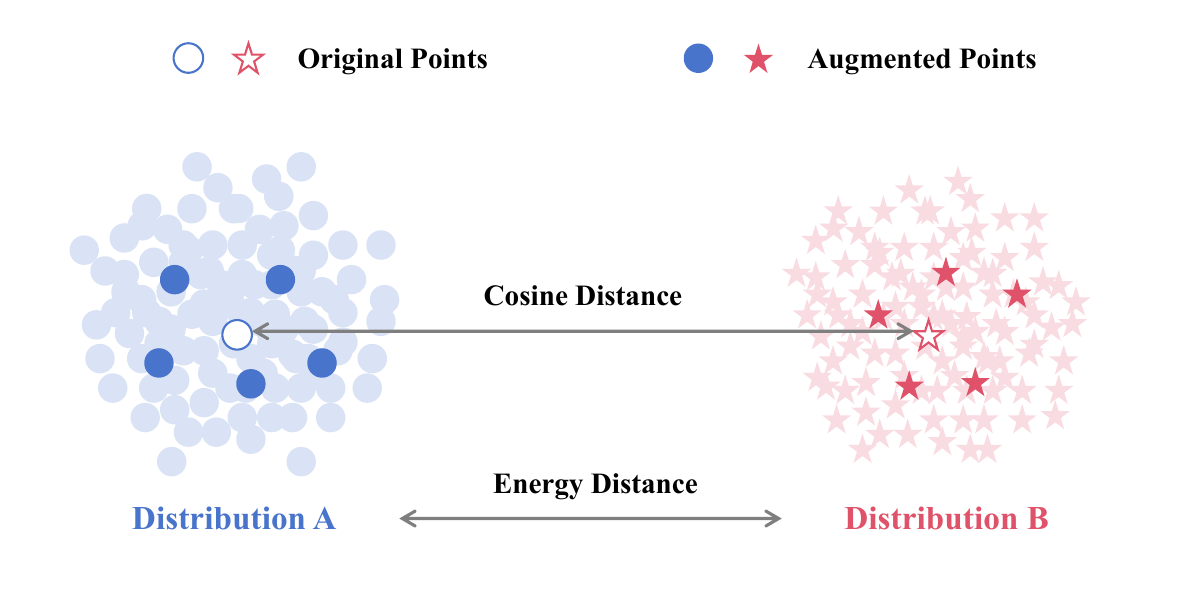}
    \caption{Comparison between cosine distance and energy distance. We leverage the randomness of data augmentation to estimate the latent semantics distributions, and then use energy distance between distributions as a substitute for cosine distance between data points.}\label{intro}
    \vspace{-3mm}
\end{figure}

Image-text matching~\cite{sun2023unifine,zhang2022negative,huang2022mack,liu2019strong,hu2023multimodal} is a pivotal task in both computer vision and natural language processing, which bridges data across heterogeneous modalities. The objective is to return images correlated with a given textual description and detect texts corresponding to a given image. Considering explosively growing web data~\cite{krotov2023big}, there is a significant demand for an efficient approach that can select a small candidate set from a comprehensive dataset. Towards this end, hashing has become prevalent in information retrieval~\cite{10.1145/3474085.3475570}, particularly image-text matching~\cite{ucch,sun2022deep,tu2023unsupervised,zeng2023learning,cao2022image}, which involves mapping both texts and images into a shared binary space (Hamming space), and then determining cross-modal similarity scores by comparing their binary codes.

In literature, numerous approaches have been developed for cross-modal hashing~\cite{jiang2017deep,kaur2021comparative}, which can broadly be categorized into supervised and unsupervised methods. Supervised methods~\cite{chen2019scratch,jia2021scaling,gu2019adversary} typically incorporate ground truth similarities into a pairwise~\cite{fan2023three} or rankwise~\cite{liu2023deep} loss objective. However, due to the high costs associated with label annotation, unsupervised approaches~\cite{tu2023unsupervised,zeng2023learning,cao2022image} tend to be more appreciated in real-world applications. Unsupervised cross-modal hashing approaches typically begin by reconstructing the similarity structure between different modalities, which subsequently provides guidance during the learning process of the hashing model.

Despite the notable advancements, prevailing unsupervised cross-modal hashing approaches~\cite{tu2023unsupervised,zeng2023learning,cao2022image} still suffer from two major limitations: (1) \textit{Biased Similarity Structure.} These approaches typically employ natural distances (e.g., cosine distance) to generate the semantic similarity structure. Since deep features with the same semantics should be from a high-dimensional distribution, utilizing cosine distance would be imprecise at the distribution boundaries, which generates noisy supervision, and serious error accumulation during subsequent optimization procedures. (2) \textit{Distribution Discrepancy Across Modalities.} Given the inherent heterogeneity, different networks are utilized to generate binary codes, which could obey distinct distributions in the Hamming space. This distribution discrepancy inherently undermines the effectiveness of cross-modal retrieval and brings suboptimal results.

To handle these limitations, in this work, we propose a new hashing approach named \underline{D}istribution-based Structur\underline{e} \underline{M}ining with C\underline{o}nsistency Learning (\method{}) for efficient image-text matching. 
The core of our \method{} revolves around exploring the latent semantic distribution of each sample using multiple random augmentations. In particular, given that data augmentation generally maintains the semantics~\cite{dai2023chataug}, we consider each augmented view of an image as samples drawn from its intrinsic semantic distribution. Then a non-parametric metric (i.e., energy distance~\cite{rizzo2016energy}) is incorporated to precisely measure the distribution divergence (see Figure \ref{intro}), thereby reconstructing a robust and accurate semantic structure. The subsequent optimization of the hashing network is achieved by preserving this semantic structure in the Hamming space. 
Furthermore, to diminish the distribution shift across modalities, we generate cross-modal retrieval distributions given both queries of images and texts and their consistency are promoted in a self-supervised manner. In addition, we employ a sharpening operation to refine retrieval results by emphasizing points with high degrees of similarity. We conduct comprehensive experiments on three benchmark image-text matching datasets, and the results show that our \method{} outperforms a wide range of competing methods. In brief, the main contribution of this paper can be summarized as follows:

\begin{itemize}[leftmargin=*]
  \item \textit{Innovative Perspective.} We explore the latent semantics distribution and adopt the distribution divergence to construct a robust and accurate semantics structure to guide unsupervised cross-modal hashing through a statistical perspective.
  \item \textit{Coherent Framework.} \method{} optimizes the modality-specific hashing networks by preserving the semantics structure in the Hamming space. Additionally, \method{} promotes consistency between cross-modal retrieval distributions, resulting in modality-invariant binary descriptors.
  \item \textit{Outstanding Performance.} Comprehensive experiments reveal that \method{} outperforms various state-of-the-art hashing-based methods on image-text matching benchmark datasets.
\end{itemize}

%% file: 3_related.tex
\section{Related Work}

\subsection{Image-text Matching}

Image-text matching is a fundamental problem which can bridge computer vision and natural language processing~\cite{sun2023unifine,zhang2022negative,huang2022mack,liu2019strong,hu2023multimodal}. Recent approaches can be divided into local-level and global-level approaches. Local-level matching approaches~\cite{liu2019focus,chen2020imram,zhang2022show,dong2022hierarchical,fu2023learning,bhattacharyya2022aligning} take the input of image-text pairs to learn fine-grained relationships, such as region-word alignments. In contrast, global-level matching approaches~\cite{tu2021hashing,lu2022cots,radford2019language,jia2021scaling} map both images and texts into a shared space and then calculate their latent embedding similarities. To enhance the efficiency of image-text matching, this paper proposes a novel hashing method termed \method{} for binary descriptors, which enables the calculation of similarity using the efficient ``XOR" operation~\cite{gu2022accelerating}.  

\begin{figure*}[t]
    \centering
    \includegraphics[width=\linewidth]{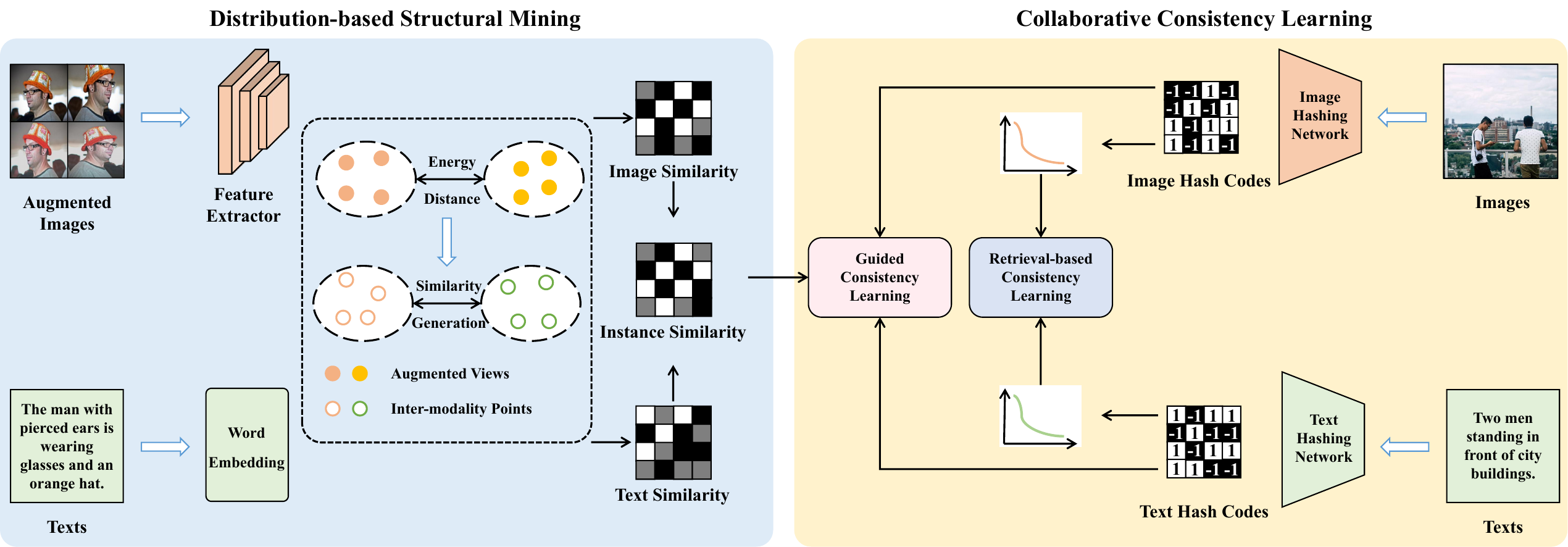}
    \caption{An overview of our proposed \method{}. \method{} first calculates the energy distance between latent semantics distributions to generate an instance similarity matrix. Then \method{} simultaneously optimizes the modality-specific hashing networks by preserving the similarity with guided consistency learning. In addition, retrieval distributions using both image and text queries are encouraged to be consistent to obtain modality-invariant binary codes. }
    \label{framework}
\end{figure*}

\subsection{Unsupervised Cross-modal Hashing}

Cross-modal hashing~\cite{ucch,sun2022deep,tu2023unsupervised,zeng2023learning,cao2022image} attempts to project samples from various modalities into a shared binary space in which samples with similar semantics should be close. Early efforts typically investigate hand-crafted features for hash codes~\cite{song2013inter,lssh}, which are typically not discriminative to preserve similarity structure. Recently, various deep unsupervised cross-modal hashing approaches have been developed~\cite{gao2023long,mikriukov2022unsupervised}, which typically reconstruct the similarity structure based on cosine distances to optimize the process of learning to hash. However, these methods are incapable of producing precise supervision signals, resulting in inferior binary hash codes. Towards this end, we investigate the latent distribution for each sample and adopt the distribution divergence for enhanced semantic structures.

%% file: 4_method.tex
\section{The Proposed Approach}

\subsection{Problem Definition and Overview}

\textbf{Problem Definition.} We begin with notations and the formal definition. $X=\{\bm{x}_i\}_{i=1}^N$ represents a dataset consisting of $N$ images and $Y= \{\bm{y}_i\}_{i=1}^N$ represent a dataset with $N$ texts. Each $\bm{y}_i$ is associated with text embeddings $\bm{t}_i$. The objective is to map these samples into a shared Hamming space. We expect the matched samples between two modalities to be encoded into similar binary codes with small Hamming distances. This mapping can guarantee effective and efficient image-text matching.

\noindent\textbf{Framework Overview.} This work proposes a new cross-modal hashing approach named \method{} for efficient image-text matching. As depicted in Figure \ref{framework}, \method{} first employs a pre-trained feature extractor $F^v(\cdot)$ for images, which removes the last layer of a well-known classification neural network~\cite{tu2023unsupervised}. We also extract text embeddings using an embedding layer $F^t(\cdot)$. Then, two feed-forward networks (FFNs), $\phi^v(\cdot)$ and $\phi^t(\cdot)$ are adopted to map features of images and texts into binary codes, respectively. Formally, we have:
\begin{equation}
    \bm{b}^v_i= sgn(\phi^v (F^v(\bm{x}_i))),
\end{equation}
\begin{equation}
    \bm{b}^t_i= sgn(\phi^t (F^t(\bm{y}_i))),
\end{equation}
where $sgn(\cdot)$ is the sign function. Our \method{} mainly consists of two modules, (1) \textit{Distribution-based Structural Mining.} We delve into the inherent semantics distribution behind each image using random data augmentation and utilize the distribution divergence to reconstruct an accurate semantic structure, which would effectively guide the optimization of hashing networks. (2) \textit{Collaborative Consistency Learning.} On the one hand, we maximize the consistency of similarity scores between the semantic structure and hash codes. On the other hand, we produce cross-modal retrieval distributions given texts and images and encourage their consistency from opposing directions.

\subsection{Distribution-based Structural Mining}

A pivotal challenge in unsupervised cross-modal hashing lies in the lack of supervised information. Previous approaches~\cite{dgcpn,uchstm,ucch} typically reconstruct the similarity structure as supervision by measuring the natural distance (e.g., cosine distance) of deep features. However, the reconstructed structure may introduce noise, leading to significant error accumulation throughout subsequent optimization stages. In particular, we observe that deep features with the same semantics should originate from a high-dimensional distribution~\cite{sun2022heart,yang2018semantic,tu2020mls3rduh}, and the natural distance could be inaccurate at the boundaries of latent distributions. Consequently, we aim to measure the distribution divergence for effective structural mining, ensuring high-quality hash codes for efficient image-text matching.

Firstly, we take the image dataset as an example of similarity structure mining. In particular, the random vector of each example $\bm{x}_i$ in the embedding space is represented as $\bm{\xi}_i$ with the cumulative distribution function $G_i$. Then, the distribution divergence between the underlying semantic distributions of $\bm{x}_i$ and $\bm{x}_j$ is formulated as:
\begin{equation}
    d(\bm{x}_i,\bm{x}_j ) = \psi(G_i, G_j ),
\end{equation}
in which $\psi$ is a given metric. However, due to immense complexity, parameterizing the high-dimensional distributions remains a considerable challenge. Therefore, classic methods such as KL divergence and JS divergence are not inappropriate here. Towards this end, we turn to a non-parametrized metric, i.e., energy distance~\cite{szekely2013energy} . This metric enables modeling of the distribution divergence without the derivation of specific distribution functions, providing an effective alternative for handling the challenges in the high-dimensional space. 

\begin{definition}
    (Energy Distance). Given two independent random vectors $\bm{\xi}$ and $\bm{\zeta}$ with the cumulative distribution functions $G_{\bm{\xi}}$ and $G_{\bm{\zeta}}$, respectively. We construct two independent copies $\bm{\xi}'$ and $\bm{\zeta}'$ from these cumulative distribution functions. Then, the energy distance is defined as:
\begin{equation}\label{energy:definition}
D^{2}(G_{\bm{\xi}}, G_{\bm{\zeta}})=2 \mathrm{E} \rho(\xi, \zeta)-\mathrm{E} \rho\left(\xi, \xi^{\prime}\right)-\mathrm{E} \rho\left(\zeta, \zeta^{\prime}\right),
\end{equation}
where $\rho(\cdot,\cdot)$ is a pointwise distance metric such as cosine distance.  
\end{definition}
When random variables are real-valued, we can rewrite Eqn. \ref{energy:definition} into:
\begin{equation}\label{eq:real-value}
D^{2}(G_{\bm{\xi}}, G_{\bm{\zeta}})=\int_{-\infty}^{\infty} \rho^{2}(G_{\bm{\xi}}(x),G_{\bm{\zeta}}(x)) d x.
\end{equation}
From Eqn. \ref{eq:real-value}, we can infer that $D^{2}(G_{\bm{\xi}}, G_{\bm{\zeta}})\geq 0$ and the equality holds when two distributions are identical. In non-parametric test, we generate statistical samples $\{\bm{u}_1,\cdots, \bm{u}_M\}$ and $\{\bm{v}_1,\cdots, \bm{v}_M\}$ from $G_{\bm{\xi}}$ and $G_{\bm{\zeta}}$, respectively. Then, we explore the statistics for the null hypothesis, i.e., $G_{\bm{\xi}}= G_{\bm{\zeta}}$ by calculating the following averages:
\begin{equation}
\begin{array}{l}A=\frac{1}{M^{2}} \sum_{m=1}^{M} \sum_{m^{\prime}=1}^{M} \rho\left(\boldsymbol{u}_{m},\boldsymbol{v}_{m^{\prime}}\right) \\ B=\frac{1}{M^{2}} \sum_{m=1}^{M} \sum_{m^{\prime}=1}^{M} \rho\left(\boldsymbol{u}_{m},\boldsymbol{u}_{m^{\prime}}\right) \\ C=\frac{1}{M^{2}} \sum_{m=1}^{M} \sum_{m^{\prime}=1}^{M} \rho\left(\boldsymbol{v}_{m},\boldsymbol{v}_{m^{\prime}}\right)\end{array}.
\end{equation}
The statistics~\cite{szekely2013energy} can be formulated as:
\begin{equation}
\mathcal{E}\left(\left\{\boldsymbol{u}_{m}\right\}_{m=1}^{M},\left\{\boldsymbol{v}_{m}\right\}_{m=1}^{M}\right)=2 A-B-C,
\end{equation}
where $\mathcal{E}(\cdot,\cdot)$ denotes energy distance. A large energy distance would reject the null hypothesis, indicating different distribution functions. Since the labels of unlabeled samples cannot acquired, we turn to data augmentation~\cite{sun2022heart,luo2020cimon,he2020momentum}. In particular, we view the augmented view of each image $\bm{x}_i$ as the samples from its underlying semantic distribution $G_i$ since data augmentation would typically retain the semantics. Therefore, the distribution divergence between $\bm{x}_i$ and $\bm{x}_j$ can be estimated as:
\begin{equation}\label{eq:distribution divergence}
d\left(\boldsymbol{x}_{i}, \boldsymbol{x}_{j}\right)=\mathcal{E}\left(\left\{\boldsymbol{z}^{\prime}{ }_{i m}\right\}_{m=1}^{M},\left\{\boldsymbol{z}^{\prime}{ }_{j m}\right\}_{m=1}^{M}\right),
\end{equation}
where $\boldsymbol{z}^{\prime}{ }_{i m} = F^v(\bm{x}'_{im} )$ is the deep feature of the augmented view $\bm{x}'_{im}$. In our implementation, we use cosine distance for $\rho(\cdot, \cdot)$. Finally, we set a threshold $\tau$ to reject the null hypothesis and thus the pair with the distance below the threshold is considered as positive. Moreover, we notice there are still fine-grained differences among dissimilar pairs. Towards this end, we introduce image and text similarities in the semantics structure:
\begin{equation}
    S_{ij}^{v}=\rho(\sum_{m=1}^M \boldsymbol{z}^{\prime}{ }_{i m},\sum_{m=1}^M \boldsymbol{z}^{\prime}{ }_{j m}),
\end{equation}
\begin{equation}
    S_{ij}^{t}=\rho(\bm{t}_i,\bm{t}_j),
\end{equation}
where $\bm{t}_i$ is the text embedding of $\bm{y}_i$. We combine $S_{ij}^{v}$ and $S_{ij}^{t}$ to depict the similarities when $d\left(\boldsymbol{x}_{i}, \boldsymbol{x}_{j}\right) \geq \tau$. In formulation, we construct the instance similarity structure as follows:
\begin{equation}\label{eq:instance_similarity}
   S_{ij}=\left\{\begin{array}{ll}
   1, & d\left(\boldsymbol{x}_{i}, \boldsymbol{x}_{j}\right)<\tau \\ 
   \alpha S_{ij}^{v}+(1-\alpha)S_{ij}^{t} , &otherwise,\end{array}\right.
    \end{equation} 
where $\alpha$ is a coefficient to balance two similarities~\cite{uchstm,tu2020mls3rduh,ma2022improved}.
It can be noticed that when $M=1$, our distribution divergence would be degraded to the fundamental cosine distance. The incorporation of multiple augmented views makes it more robust against random attacks. Moreover, it alleviates biases for examples at the boundary of latent semantic distributions, ensuring the accuracy of structural mining. 

\subsection{Optimization with Collaborative Consistency Learning}

In this part, we jointly optimize the image and text hashing networks using collaborative consistency learning which mainly includes guided consistency learning and retrieval-based consistency learning. 

\noindent\textbf{Guided Consistency Learning.} After constructing the similarity structure~\cite{luo2020cimon,yang2018semantic,tu2020mls3rduh}, we aim to preserve this in produced hash codes. In particular, we generate hash codes for both images and texts and then produce their similarities, which would be consistent with the reconstructed structure. Formally, we have:
\begin{equation}
    \mathcal{L}_{gui} = \sum_{i,j=1}^N \sum_{e_1,e_2\in\{v,t\}}||\rho(\bm{b}_i^{e_1},\bm{b}_j^{e_2})-S_{ij}||^2,
\end{equation}
where $e_1$ and $e_2$ indicate the selected modalities. Therefore, image-image, text-text, and image-text consistency are jointly considered and mapped to the similarity structure under the guidance. 

\noindent\textbf{Retrieval-based Consistency Learning.} To further reduce the potential distribution discrepancy between the two modalities~\cite{lu2022cots,wei2020co2}, we simulate the cross-modal retrieval procedure in different directions and enforce the consistency between the retrieval results. In formulation, given a batch, the probability distribution corresponding to text-to-image retrieval is written as:
\begin{equation}\label{eq:t2i}
    \bm{p}^{T2I}_i = [\rho(\bm{b}_i^t,\bm{b}_1^v),\cdots,\rho(\bm{b}_i^t,\bm{b}_B^v)],
\end{equation}
where $B$ denotes the batch size. Similarly, the probability distribution corresponding to image-to-text retrieval is:
\begin{equation}\label{eq:i2t}
    \bm{p}^{I2T}_i = [\rho(\bm{b}_i^v,\bm{b}_1^t),\cdots,\rho(\bm{b}_i^v,\bm{b}_B^t)].
\end{equation}
Then, we utilize the sharpening operator~\cite{xie2016unsupervised,assran2021semi,wang2023dior} to refine the soft distributions with:
\begin{equation}
   \delta(\bm{p})_b = \frac{[\bm{p}]_b^{1/T}}{\sum_{b'=1}^{B}[\bm{p}]_{b'}^{1/T}}, b=1,\cdots,B. 
\end{equation}
Our sharpening operation is capable of enhancing the purification of the retrieval results and emphasizing the samples with high similarities. Finally, we conduct consistency learning across two directions in a self-supervised fashion using:
\begin{equation}
\begin{aligned}
        \mathcal{L}_{ret} &= \sum_{i=1}^B (KL(\delta(\bm{p}^{I2T}_i)||\bm{p}^{T2I} ) \\&+ KL(\delta(\bm{p}^{T2I}_i)||\bm{p}^{I2T} )),
\end{aligned}
\end{equation}
where $KL(\cdot||\cdot)$ returns the KL divergence of two distributions and $T$ is a temperature coefficient that controls the sharp degree set to $0.25$ empirically. 

Besides, we leverage the co-occurrence knowledge embedded in the dataset, which enforces binary codes of images and texts with identical objects to be close. In particular, we have:
\begin{equation}
    \mathcal{L}_{co} = \sum_{i=1}^N   ||\rho(\bm{b}_i^{v},\bm{b}_j^{t})-\gamma||^2,
\end{equation}
where $\gamma$ is set to $1.5$ empirically~\cite{tu2023unsupervised} to emphasize this accurate embedding knowledge. 

In a nutshell, we summarize our framework by combining all these objectives:
\begin{equation}\label{eq:loss}
    \mathcal{L} = \mathcal{L}_{gui} + \mathcal{L}_{ret} + \mathcal{L}_{co}.
\end{equation}
However, directly minimizing Eqn. \ref{eq:loss} is infeasible since $sgn(\cdot)$ is not differentiable at zero and its derivative is zero at the other point. To tackle this problem, we replace $sgn(\cdot)$ with $tanh(\cdot)$ during optimization, which results in approximate hash codes $\hat{\bm{b}}^v_i = tanh(\phi^v (F^v(\bm{x}_i)))$ and $\hat{\bm{b}}^v_t = tanh(\phi^t (F^t(\bm{y}_i)))$. 
We summarize the whole training algorithm of \method{} in Algorithm \ref{alg1}.

\subsection{Model Inference}

After the optimization procedure, we would feed each sample into the hashing network for a binary descriptor. Then, given each query text $y_q$ (image $x_q$) with a binary code $\bm{b}_q^t$ ($\bm{b}_q^v$), we rank the Hamming distances between $\bm{b}_q^t$ ($\bm{b}_q^v$) and $\{\bm{b}_i^v\}_{i=1}^N$ ($\{\bm{b}_i^t\}_{i=1}^N$), which can produce the nearest examples efficiently. In practice, we consider the returned samples as candidates and conduct fine-grained matching for the final results~\cite{tu2021hashing}.

%% file: 5_experiment.tex
\section{Experiment}
\begin{table*}[t]
\centering
\resizebox{\linewidth}{!}{%
\begin{tabular}{c|c|cccc|cccc|cccc}
\toprule
\multirow{2}{*}{\textbf{Task}} & \multirow{2}{*}{\textbf{Method}} & \multicolumn{4}{c|}{\textbf{MIRFlickr-25K}} & \multicolumn{4}{c|}{\textbf{NUS-WIDE}} & \multicolumn{4}{c}{\textbf{MS-COCO}} \\
 &  & \textbf{16 bits} & \textbf{32 bits} & \textbf{64 bits} & \textbf{128 bits} & \textbf{16 bits} & \textbf{32 bits} & \textbf{64 bits} & \textbf{128 bits} & \textbf{16 bits} & \textbf{32 bits} & \textbf{64 bits} & \textbf{128 bits} \\
\midrule 
\multirow{11}{*}{\textbf{I2T}} & CVH & 0.620 & 0.608 & 0.594 & 0.583 & 0.487 & 0.495 & 0.456 & 0.419 & 0.503 & 0.504 & 0.471 & 0.425 \\
 & LSSH & 0.597 & 0.609 & 0.606 & 0.605 & 0.442 & 0.457 & 0.450 & 0.451 & 0.484 & 0.525 & 0.542 & 0.551 \\
 & CMFH & 0.557 & 0.557 & 0.556 & 0.557 & 0.339 & 0.338 & 0.343 & 0.339 & 0.366 & 0.369 & 0.370 & 0.365 \\
 & FSH & 0.581 & 0.612 & 0.635 & 0.662 & 0.557 & 0.565 & 0.598 & 0.635 & 0.539 & 0.549 & 0.576 & 0.587 \\
 & MTFH & 0.507 & 0.512 & 0.558 & 0.554 & 0.297 & 0.297 & 0.272 & 0.328 & 0.399 & 0.293 & 0.295 & 0.395 \\
 & FOMH & 0.575 & 0.640 & 0.691 & 0.659 & 0.305 & 0.305 & 0.306 & 0.314 & 0.378 & 0.514 & 0.571 & 0.601 \\
 & DCH & 0.596 & 0.602 & 0.626 & 0.636 & 0.392 & 0.422 & 0.430 & 0.436 & 0.422 & 0.420 & 0.446 & 0.468 \\
 & DGCPN & 0.651 & 0.683 & 0.718 & 0.724 & 0.601 & 0.618 & 0.631 & 0.640 & 0.556 & 0.569 & 0.578 & 0.580 \\
 & UCHSTM & 0.701 & 0.715 & 0.724 & 0.723 & 0.625 & 0.635 & 0.646 & 0.644 & 0.558 & 0.572 & 0.576 & 0.573 \\
 & UCCH & 0.716 & 0.726 & 0.728 & 0.732 & 0.621 & 0.623 & 0.640 & 0.645 & 0.560 & 0.562 & 0.566 & 0.574 \\
 & \cellcolor{blue!20} \method{} & \cellcolor{blue!20} \textbf{0.718} &\cellcolor{blue!20} \textbf{0.733} &\cellcolor{blue!20} \textbf{0.734} &\cellcolor{blue!20} \textbf{0.743} &\cellcolor{blue!20} \textbf{0.646} &\cellcolor{blue!20} \textbf{0.648} &\cellcolor{blue!20} \textbf{0.662} &\cellcolor{blue!20} \textbf{0.664} &\cellcolor{blue!20} \textbf{0.575} &\cellcolor{blue!20} \textbf{0.578} &\cellcolor{blue!20} \textbf{0.586} &\cellcolor{blue!20} \textbf{0.605} \\
\midrule
\multirow{11}{*}{\textbf{T2I}} & CVH & 0.629 & 0.615 & 0.599 & 0.587 & 0.470 & 0.475 & 0.444 & 0.412 & 0.506 & 0.508 & 0.486 & 0.429 \\
 & LSSH & 0.602 & 0.598 & 0.598 & 0.597 & 0.473 & 0.482 & 0.471 & 0.457 & 0.490 & 0.522 & 0.547 & 0.560 \\
 & CMFH & 0.553 & 0.553 & 0.553 & 0.553 & 0.306 & 0.306 & 0.306 & 0.306 & 0.346 & 0.346 & 0.346 & 0.346 \\
 & FSH & 0.576 & 0.607 & 0.635 & 0.660 & 0.569 & 0.604 & 0.651 & 0.666 & 0.537 & 0.524 & 0.564 & 0.573 \\
 & MTFH & 0.514 & 0.524 & 0.518 & 0.581 & 0.353 & 0.314 & 0.399 & 0.410 & 0.335 & 0.374 & 0.300 & 0.334 \\
 & FOMH & 0.585 & 0.648 & 0.719 & 0.688 & 0.302 & 0.304 & 0.300 & 0.306 & 0.368 & 0.484 & 0.559 & 0.595 \\
 & DCH & 0.612 & 0.623 & 0.653 & 0.665 & 0.379 & 0.432 & 0.444 & 0.459 & 0.421 & 0.428 & 0.454 & 0.471 \\
 & DGCPN & 0.653 & 0.682 & 0.712 & 0.715 & 0.605 & 0.626 & 0.637 & 0.644 & 0.550 & 0.566 & 0.578 & 0.577 \\
 & UCHSTM & 0.695 & 0.711 & 0.713 & 0.723 & 0.632 & 0.643 & 0.651 & 0.652 & 0.555 & 0.567 & 0.578 & 0.573 \\
 & UCCH & 0.703 & 0.712 & 0.720 & 0.721 & 0.625 & 0.637 & 0.650 & 0.652 & 0.564 & 0.573 & 0.572 & 0.581 \\
 & \cellcolor{blue!20}\method{} & \cellcolor{blue!20}\textbf{0.708} & \cellcolor{blue!20}\textbf{0.719} &\cellcolor{blue!20} \textbf{0.722} & \cellcolor{blue!20}\textbf{0.728} & \cellcolor{blue!20}\textbf{0.654} &\cellcolor{blue!20} \textbf{0.655} &\cellcolor{blue!20} \textbf{0.669} & \cellcolor{blue!20}\textbf{0.671} & \cellcolor{blue!20}\textbf{0.572} & \cellcolor{blue!20}\textbf{0.579} &\cellcolor{blue!20} \textbf{0.583} & \cellcolor{blue!20}\textbf{0.597} \\
\bottomrule
\end{tabular}%
}
\caption{MAP scores comparison with code length varying from 16 to 128 bits. I2T refers to the image-to-text matching task, and T2I signifies the text-to-image task. The highest scores are shown in \textbf{boldface}.}
\label{main res}
\end{table*}

\begin{figure*}[t]
    \centering
    \includegraphics[width=\linewidth]{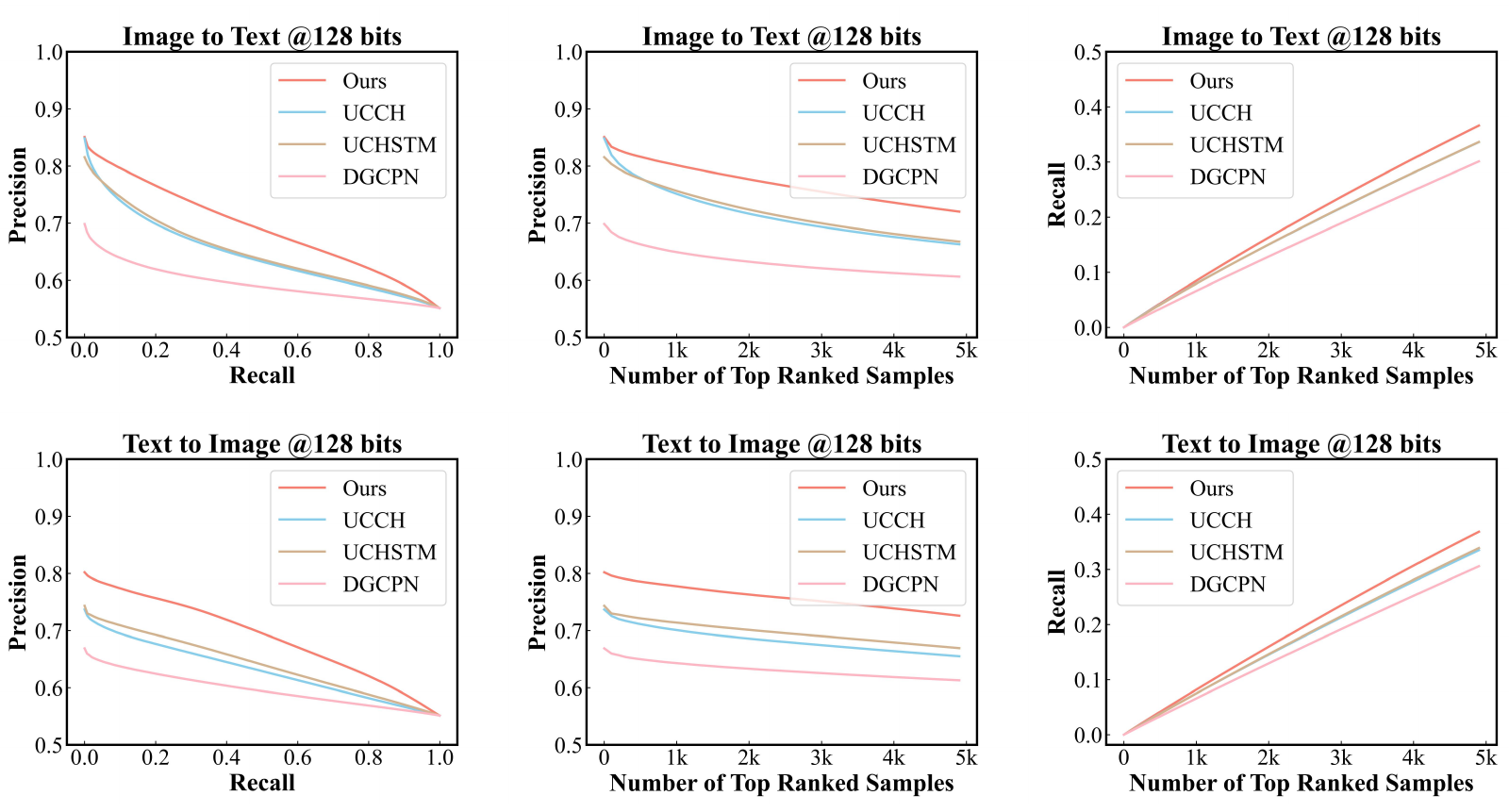}
    \caption{The Precision-Recall curve,  Precision-top N curve, and Recall-top N curve with 128 bits on MIRFlickr-25K. The first row plots image-to-text results, and the second row plots text-to-image results.}
    \label{curve_128}
    \vspace{-0.2cm}
\end{figure*}

\subsection{Experimental Settings}
\noindent\textbf{Datasets and Evaluation Metrics.}
To assess the performance of our \method{}, we employ three public and widely-used benchmark datasets to conduct experiments, including MIRFlickr-25K, NUS-WIDE, and MS-COCO.
\textit{MIRFlickr-25K} comprises 25,000 pairs of image-text data, and each sample is manually annotated with multiple labels from a set of 24 distinct classes. We remove samples lacking class information, resulting in 20,015 samples for our experiments. We divide these samples into two sets: a query database containing 2,000 paired samples and a retrieval database containing the remaining samples. We employ bag-of-words (BoW) vectors with a dimension of 1,386 to represent the text samples.
\textit{NUS-WIDE} consists of 269,498 paired image-text samples and each sample is assigned to a multilabel category from 81 categories. We select 186,557 samples from the top 10 frequent classes for our experiments. These samples are split into a query database with 2,100 image-text pairs and a retrieval database with the remaining samples. Similarly, we employ 1,000-dimensional BoW vectors to represent the text samples.
\textit{MS-COCO} is a benchmark dataset which consists of 123,287 images. Each image is associated with 5 annotations from 80 categories. After deleting the samples without label annotations, 122,218 pairs remain during the experiment. We choose 5,000 paired image-text samples randomly as the query database and the remaining pairs are left as the retrieval database. Correspondingly, the text samples are represented by 2026-dimensional BoW vectors.

We evaluate the matching performance based on two protocols: the Hamming ranking protocol and the hash lookup protocol. The former is evaluated by the widely used metric Mean Average Precision (MAP) score, and the latter is evaluated by three types of curves: Precision-Recision curve, Precision-top N curve, and Recall-top N curve. For a fair comparison, we report MAP@All scores as default.

\noindent\textbf{Baselines and Implementation Details.}
We employ 10 state-of-the-art hashing-based image-text matching approaches as baseline methods, including three supervised cross-modal hashing methods (MTFH~\cite{mtfh}, FOMH~\cite{fomh}, DCH~\cite{dch}), four shallow unsupervised cross-modal hashing methods (CVH~\cite{cvh}, LSSH~\cite{lssh}, CMFH~\cite{cmfh}, FSH~\cite{fsh}), and three deep unsupervised cross-modal hashing methods (DGCPN~\cite{dgcpn}, UCHSTM~\cite{uchstm}, UCCH~\cite{ucch}). 
We randomly select 5,000/10,000/all samples from the retrieval database as the training samples for supervised/deep unsupervised/shallow unsupervised cross-modal hashing methods. For a fair comparison, we follow previous works~\cite{uchstm,ucch} and reimplement the deep unsupervised methods, utilizing VGG-19 pre-trained on the ImageNet dataset and a two-layer MLP as the backbone of the image hashing network and text hashing network, respectively. We adopt the SGD algorithm with a learning rate of 1e-3 to optimize the networks. The batch size is set to 128. More hyper-parameters are set according to Section~\ref{sec: sensitivity}.

\subsection{Main Results}
\noindent\textbf{Hamming Ranking Protocol.}
We showcase the MAP scores of all compared baseline methods and our \method{} in Table~\ref{main res}. From these results, the following observations can be attained: \textit{First}, deep unsupervised cross-modal hashing methods outperform shallow unsupervised cross-modal hashing approaches even with insufficient amounts of training data, indicating the superiority of deep neural networks in generating high-quality and modality-invariant hash codes.
\textit{Next}, supervised methods excel due to their reliance on expensive labeled data. However, when labeled data is scarce, these methods fall short compared to deep unsupervised approaches. Consequently, deep unsupervised cross-modal hashing emerges as the fundamental technique for image-text matching in the presence of vast amounts of unlabeled multimodal data. 
\textit{Furthermore}, \method{} outperforms all the compared state-of-the-art hashing-based image-text matching methods, revealing the effectiveness of our proposed distribution-based structural mining and collaborative consistency learning. Additionally, our approach exhibits consistent and significant improvements across three datasets, highlighting the success of addressing previously overlooked distribution divergence combined with collaborative consistency. The proposed components can enhance the performance of unsupervised hashing-based image-text matching in a robust manner.

\noindent\textbf{Hash Lookup Protocol.}
We also incorporate the hash lookup protocol to generate Precision-Recall, Precision-top N, and Recall-top N curves for our \method{} and three reproduced deep unsupervised baselines using 128 bits on MirFlickr-25K, as illustrated in Figure~\ref{curve_128}. 
Due to space limitations, curves for other code lengths can be seen in Section~\ref{lab:detail}. 
The Precision-Recall curve represents the relationship between the varying precision and recall scores. The Precision-top N and Recall-top N curves depict precision and recall values as the retrieval numbers vary from $1$ to $5,000$ with a step size of $100$. In brief, for these three types of curves, the higher-performing method's curve is usually above the curves of other methods. 
These curves clearly illustrate that our \method{} consistently outperforms the other baselines, underscoring its superiority. The hash lookup results are consistent with the Hamming ranking results, further validating the exceptional performance and robustness of our \method{} in image-text matching.
\begin{table}[t]
\centering
\resizebox{\linewidth}{!}{%
\begin{tabular}{c|c|c|c|c}
\toprule
\textbf{Task} & \textbf{Method} & \textbf{MIRF-25K} & \textbf{NUS-WIDE} & \textbf{MS-COCO} \\
\midrule
\multirow{4}{*}{\textbf{I2T}} & \method{} w/o D & 0.698 & 0.627 & 0.560 \\
 & \method{} w/o R & 0.705 & 0.632 & 0.565 \\
 & \method{} w/o S & 0.712 & 0.636 & 0.571 \\
 & Full Model & \textbf{0.718} & \textbf{0.646} & \textbf{0.575} \\
 \midrule
\multirow{4}{*}{\textbf{T2I}} & \method{} w/o D & 0.696 & 0.630 & 0.559 \\
 & \method{} w/o R & 0.695 & 0.634 & 0.564 \\
 & \method{} w/o S & 0.699 & 0.642 & 0.569 \\
 & Full Model & \textbf{0.708} & \textbf{0.654} & \textbf{0.572}\\
 \bottomrule
\end{tabular}
}
\caption{Ablation on each proposed component. The highest MAP scores are shown in \textbf{boldface}.}
\label{ablation}    
\vspace{-0.3cm}
\end{table}

\begin{figure}[t]
    \centering
\includegraphics[width=\linewidth]{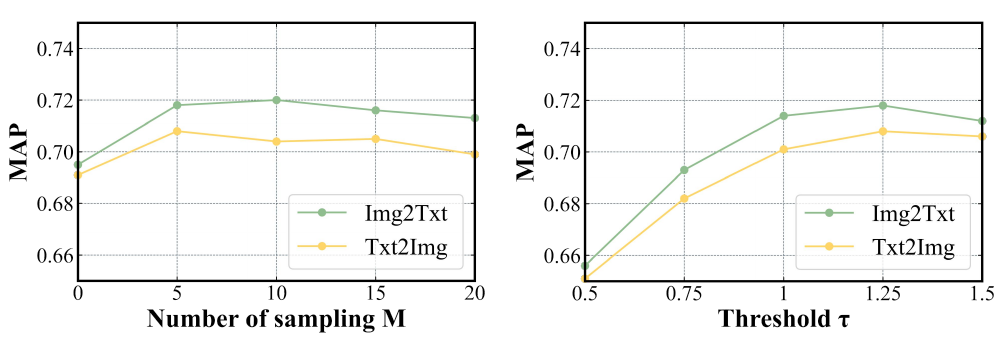}
    \caption{Sensitivity analysis of sampling times $M$ and threshold $\tau$ with 16 bits on MIRFlickr-25K.}
    \label{sensitivity}
    \vspace{-0.3cm}
\end{figure}

\begin{table}[t]
    \centering
    \resizebox{\linewidth}{!}{%
    \begin{tabular}{c|cccc}
        \toprule
        \textbf{Method} & LSSH  & UGACH & UCCH & \method{}\\
        \midrule
        \textbf{Inference Time} & 7.78s & 26.59s & 0.41s & 0.41s\\
        \bottomrule
    \end{tabular}
    }
    \caption{Comparison with other methods on the inference speed.}
    \label{tab:inference}
        \vspace{-0.3cm}
\end{table}
\begin{figure*}[t]
    \centering
    \includegraphics[width=\linewidth]{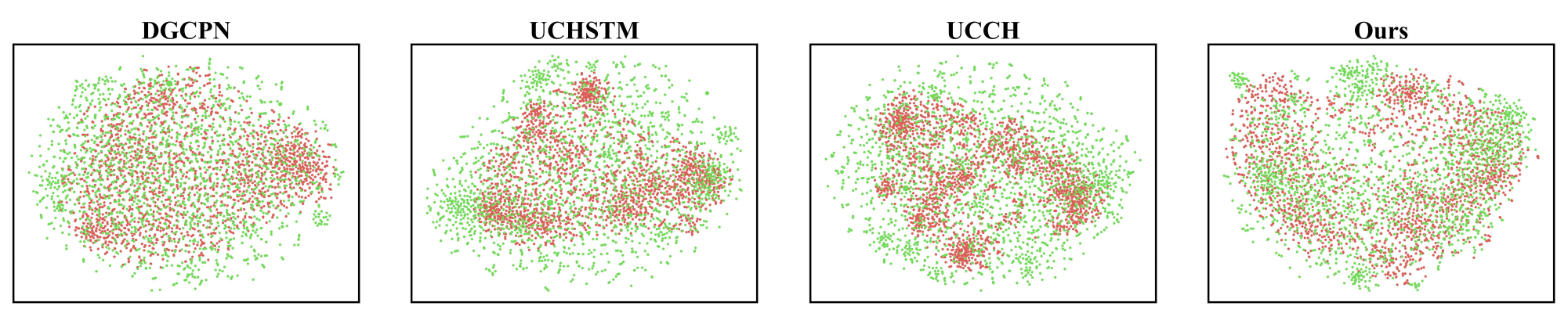}
    \caption{The t-SNE visualization with 128 bits on the MIRFlickr-25K. The image modality is colored red, and the text modality is colored green. The overlap degree represents the degree of modality-invariant hash codes.}
    \label{figure-tsne}
\end{figure*}
\label{discussion}
\subsection{Ablation Study}
In Table~\ref{ablation}, we investigate the contributions of each proposed component with 16 bits on three datasets.
\textit{Firstly}, we remove the distribution-based structural mining process and replace it with sample-based structural mining. The comparison between \method{} w/o D in the first row and the full model in the last row highlights the significant improvement achieved by our distribution-based objective. 
\textit{Next}, we assess the significance of the retrieval-based consistency learning by removing it. Without this module, the retrieved distributions given images and texts as queries are not encouraged to be consistent. The performance degradation observed in \method{} w/o R in the second row emphasizes the effectiveness of this component. 
\textit{Moreover}, we conduct an experiment where we remove only the proposed sharpening operation from the retrieval-based consistency learning module. The results of \method{} w/o S in the third row reveal slight differences compared to the full model, underscoring the importance of the sharpening operation. Furthermore, the performance of \method{} w/o S falls between \method{} w/o R and the full model, which is reasonable since \method{} w/o S removes only parts of the retrieval-based consistency learning module while still retaining the ability to promote consistency between the retrieved distributions.
\textit{Finally}, we evaluate the full model which incorporates all the components. Results in the last row exhibit the best performance across all scenarios. These experiments successfully verify the significance of each proposed component in \method{}.

\subsection{Sensitivity Analysis}
\label{sec: sensitivity}
To assess the impact of the hyper-parameter $M$ and $\tau$, Figure~\ref{sensitivity} plots the MAP scores with respect to $M$ ranging from $0$ to $20$, and $\tau$ ranging from $0.5$ to $1.5$. From the results, we can observe that increasing $M$ from $0$ to $5$ yields a significant performance improvement, but further increasing $M$ from $5$ to $20$ does not lead to any noticeable enhancement. This phenomenon demonstrates that our \method{} is not sensitive to $M$ within the range of $[5,20]$. Therefore, $M$ is fixed at $5$ and we proceed to investigate the threshold $\tau$, varying from $0.5$ to $1.5$. The threshold $\tau$ plays a crucial role as it controls the percentage of the image-text pairs categorized as positive samples, thereby influencing the quality of the generated similarity matrix. A large value of $\tau$ will mistakenly consider numerous incorrect image-text pairs as the matching ones, while a small value of $\tau$ will classify many matching image-text pairs as non-matching pairs. From the results, it can be found that $1.25$ is the most suitable for the threshold $\tau$. Consequently, we obtain the optimal value of $M=5$ and $\tau=1.25$, respectively.

\subsection{Efficiency Analysis}
We make experimental verification on the inference speed. In particular, we compare our \method{} with state-of-the-art hashing-based image-text matching approaches with 128 bits on MIRFlickr-25K. As shown in Table \ref{tab:inference}, our \method{} can achieve much higher efficiency compared with LSSH \cite{lssh} and UGACH \cite{zhang2018unsupervised}. Even though the inference time of UCCH \cite{ucch} and our \method{} is the same, our retrieval performance is much better. In summary, our \method{} is superior to these baselines taking into both efficiency and effectiveness.

\subsection{Visualization}
We present the t-SNE~\cite{tsne} visualization of hash codes from two different modalities generated by four methods with 128 bits on MirFlickr-25K in Figure~\ref{figure-tsne}. The results of the comparison with the other three approaches reveal that our \method{} demonstrates a significantly higher degree of similarity and overlap between the image and text modalities. This observation serves as a strong indication that combined with collaborative consistency learning, distribution-based structural mining is superior to sample-based structural mining. The visualization results also provide compelling evidence of the exceptional quality and modality-invariant hash codes learned by our approach.

%% file: 6_conclusion.tex
\section{Conclusion}

In this paper, we investigate the problem of image-text matching and propose a novel deep unsupervised hashing-based approach termed \method{}. The crux of our \method{} is to explore the latent semantic distributions of each sample for effective semantics structure mining. Specifically, we characterize each image with multiple augmented views, which are regarded as samples from its intrinsic semantic distribution. Then, a non-parametric distribution divergence is employed to ensure a robust and accurate similarity structure in the process of similarity generation, which serves as guidance for the optimization. Extensive experimental results across multiple datasets substantiate the efficacy of \method{}.

\section{Limitation}

Although our \method{} achieves promising results, it still has some limitations. First, there could be different complicated scenarios in real-world applications such as data contamination and domain shift. We would extend our \method{} to more generalization scenarios in our future works. Second, our unsupervised hashing approach \method{} targets at coarse-level retrieval. How to improve unsupervised cross-modal hashing for fine-grained cross-modal retrieval remains an open problem.

%% file: 7_appendix.tex
\appendix
\label{appendix}
\begin{figure*}[t]
    \centering
    \includegraphics[width=\linewidth]{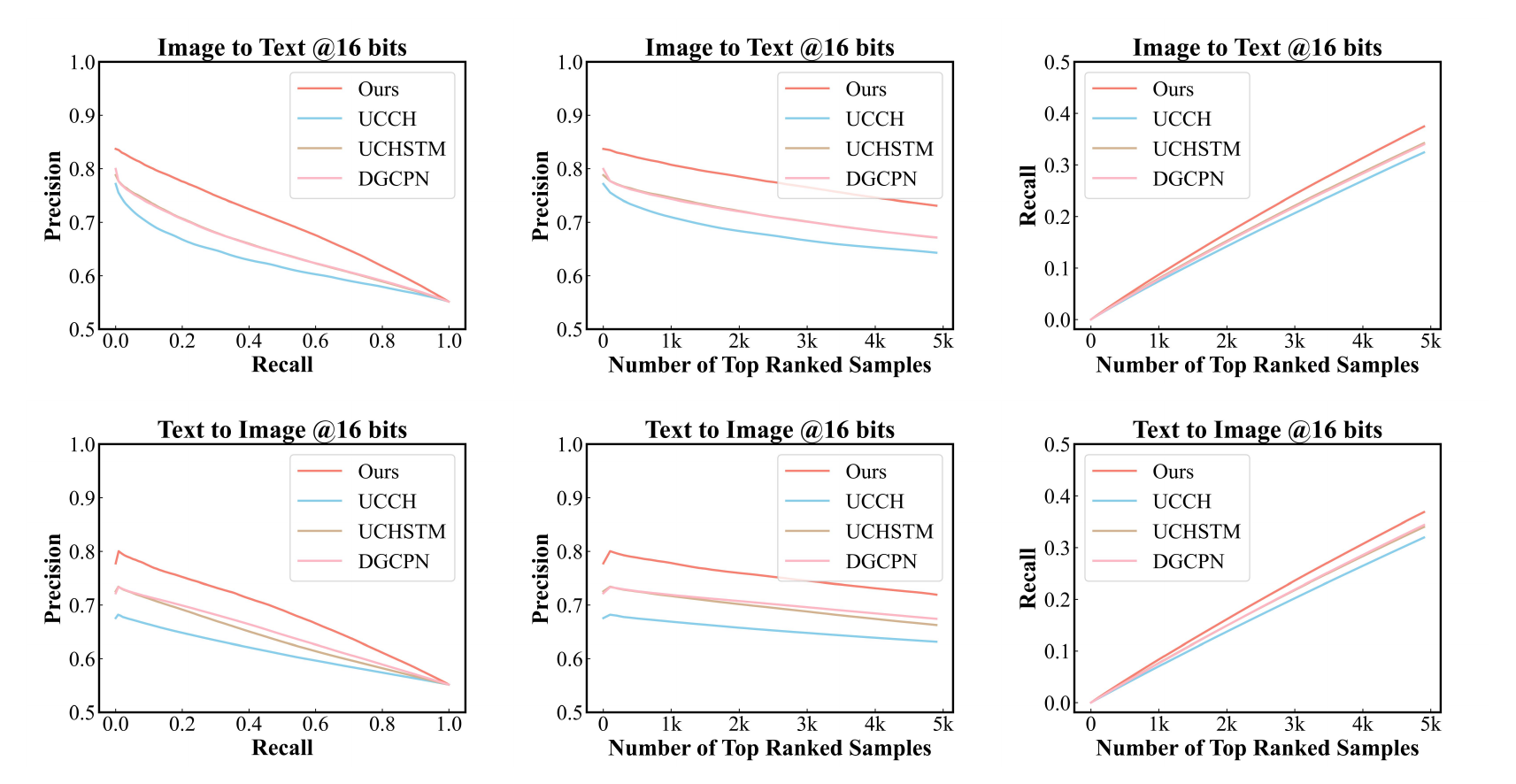}
    \caption{The Precision-Recall curve,  Precision-top N curve, and Recall-top N curve with 16 bits on the MIRFlickr-25K dataset. Image-to-text results are plotted in the first row, and text-to-image results are plotted in the second row.}
    \label{curve_16}
\end{figure*}

\begin{figure*}[t]
    \centering
    \includegraphics[width=\linewidth]{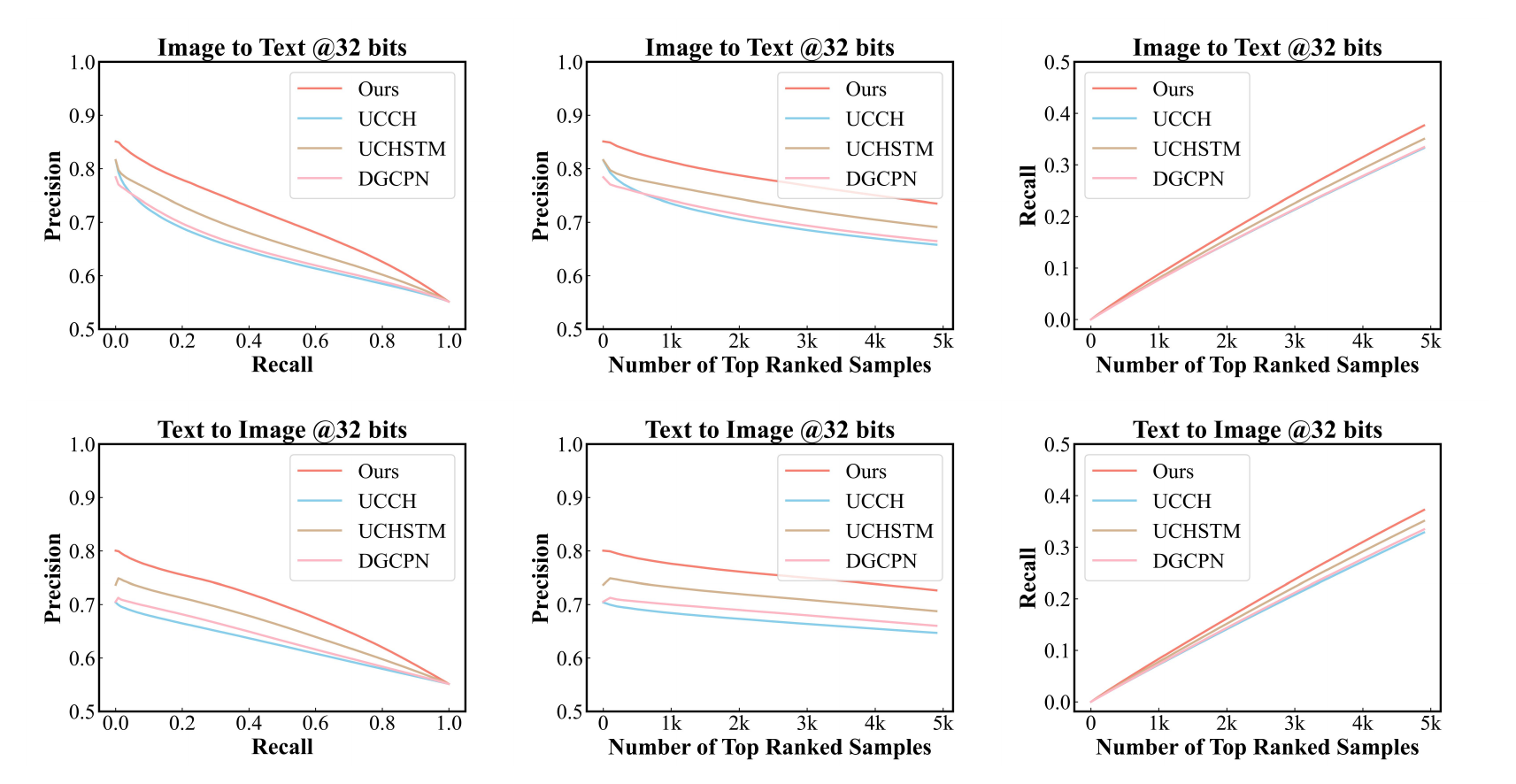}
    \caption{The Precision-Recall curve,  Precision-top N curve, and Recall-top N curve with 32 bits on the MIRFlickr-25K dataset. Image-to-text results are plotted in the first row, and text-to-image results are plotted in the second row.}
    \label{curve_32}
\end{figure*}

\begin{figure*}[!t]
    \centering
    \includegraphics[width=\linewidth]{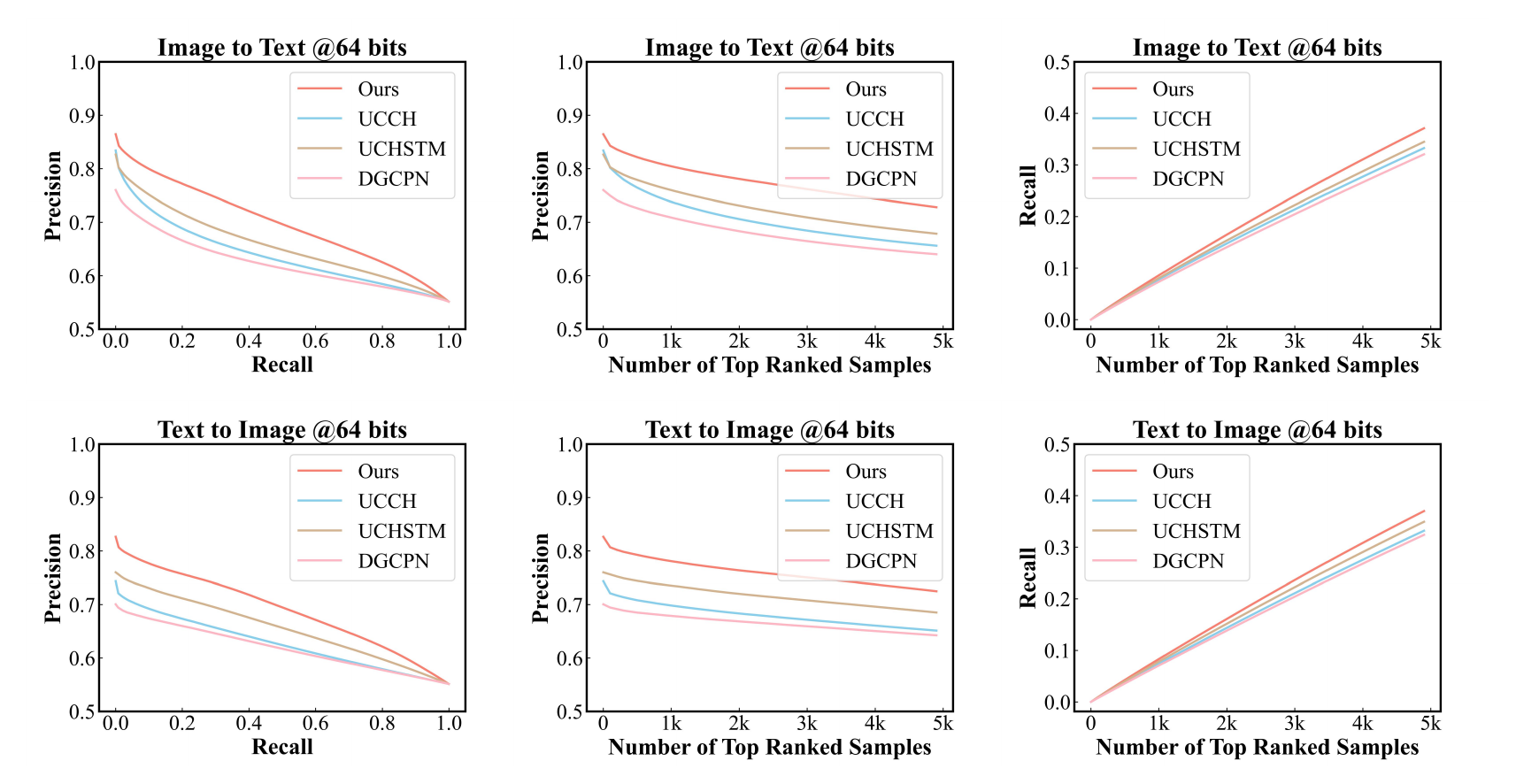}
    \caption{The Precision-Recall curve,  Precision-top N curve, and Recall-top N curve with 64 bits on the MIRFlickr-25K dataset. Image-to-text results are plotted in the first row, and text-to-image results are plotted in the second row.}
    \label{curve_64}
\end{figure*}

\section{Algorithm}
\begin{algorithm}[!h]
\caption{Training Algorithm of \method{}}
\label{alg1}
\begin{algorithmic}[1]
\REQUIRE Image dataset $X$; text dataset $Y$; number of augmented views $M$, threshold $\tau$.\\
\ENSURE Parameters of the hashing network.
\STATE Generate $M$ augmented views for every $\bm{x}_i$;
\STATE Calculate the distribution divergence using Eqn. \ref{eq:distribution divergence};
\STATE Generate the instance similarity structure using Eqn. \ref{eq:instance_similarity};
\REPEAT
\STATE Sample a mini-batch randomly;
\STATE Output approximate binary codes for both images and texts;
\STATE Generate cross-modal retrieval results using Eqn. \ref{eq:t2i} and Eqn. \ref{eq:i2t};
\STATE Calculate the whole loss using Eqn. \ref{eq:loss};
\STATE Update the hashing network by backpropagation;
\UNTIL convergence
\end{algorithmic}
\end{algorithm}

\section{Data Augmentation Strategy}
We leverage the randomness of data augmentations to convert sample-based structural mining to distribution-based structural mining. The detailed data augmentation strategy is illustrated below. First, we resize the image to $256\times256$ and randomly crop a size of $224\times224$. Then we employ strategies such as Random Horizontal Flip, Random Color Jitter with $p=0.7$, Random Grayscale with $p=0.2$, and Gaussian Blur with $kernel size=3$. Finally, we normalize the data with pre-computed mean and standard values. With this augmentation strategy, the intrinsic semantic distribution of a data sample is established for future semantic structure mining.



\section{Compared Methods}
Many state-of-the-art cross-modal hashing-based methods are employed for comparison, including three supervised methods, four shallow unsupervised methods, and three deep unsupervised methods. The detailed introduction of these methods is as follows:
\begin{itemize}
    \item \textbf{CVH}~\cite{cvh} introduces a novel relaxation technique that transforms the learning-to-hash process into a tractable eigenvalue problem. To address this challenge, they utilize techniques such as Locality Sensitive Indexing and Canonical Correlation Analysis.
    \item \textbf{LSSH}~\cite{lssh} extracts the latent semantics from textual samples by matrix factorization. It also leverages sparse coding techniques to capture essential image structures. Introducing an effective iterative method, it analyzes the correlation between multimodal representations, thereby narrowing the semantic gap within the latent semantic space. 
    \item \textbf{CMFH}~\cite{cmfh} builds robust connections via cross-modal factorization, integrating locally linear embedding to uphold the Euclidean structure. Additionally, it employs a classifier-like loss function to leverage semantic label information effectively.
    \item \textbf{FSH}~\cite{fsh} defines the similarity between different modalities by introducing a graph-based framework, and then utilizing it to learn modality-invariant hash codes.
    \item \textbf{MTFH}~\cite{mtfh} proposes to learn semantic correlations between modalities and aligns heterogeneous data to obtain modality-specific hash codes. 
    \item \textbf{FOMH}~\cite{fomh} introduces a multimodal fusion framework to fuse representations when modalities are missing, and then constructs discriminative hash codes.
    \item \textbf{DCH}~\cite{dch} optimizes the network to get modality-specific and modality-invariant hash codes simultaneously. Moreover, it refines the hash codes by iterative training to enhance efficiency.
    \item \textbf{DGCPN}~\cite{dgcpn} investigates the correlations between data samples and their neighbors to improve the quality of similarity generation. It employs a hybrid optimization strategy, combining real and binary components, to minimize discrepancies between the Hamming space and the continuous latent space, thus enhancing similarity and value consistency.
    \item \textbf{UCHSTM}~\cite{uchstm} explores correlations among words in textual data points, facilitating the creation of a text modality-specific similarity matrix derived from these correlations. Furthermore, it introduces a self-redefined similarity loss to rectify inaccuracies in the instance similarity matrix, thereby improving the accuracy of similarity measurements.
    \item \textbf{UCCH}~\cite{ucch} introduces contrastive learning, aiming to align various modalities with unified binary representations. It emphasizes leveraging discrimination from all pairs rather than solely focusing on the hardest negative pairs.
\end{itemize}

\section{Detailed Hash Lookup Protocol}
\label{lab:detail}
We showcase the hash lookup results on MIRFlickr-25K with varying code lengths in Figure~\ref{curve_16}, Figure~\ref{curve_32}, and Figure~\ref{curve_64}. From the results of the hash lookup protocol, several conclusions can be observed: 
\begin{enumerate}
    \item Firstly, from the Precision-Recall curve, we can notice the correlation between precision and recall scores. These two metrics are contradictory to each other, as an increase in one often leads to a decrease in the other. From the results in the first column, it can be found that as the recall score increases, the precision score of our \method{} consistently surpasses the other three compared baseline hashing-based image-text matching methods.
    \item Secondly, the Precision-top N curve represents the correlation between the precision score and the top N number of results returned in a single retrieval process. As the number of retrieved samples increases, the precision score tends to decrease. From the results in the second column, it can be observed that as N increases from $1$ to $5000$, our \method{} consistently outperforms the other three methods.
    \item Furthermore, similar to the Precision-top N curve, the Recall-top N curve represents the correlation between the recall score and the top N results returned in a single retrieval. Different from the precision score, the recall score tends to increase as N increases. From the results in the third column, it can be seen that as N increases from $1$ to $5000$, our curve consistently remains above the other three curves.
    \item Lastly, a large number of results demonstrate the robustness of our method from different perspectives. Whether it pertains to various code lengths or different modalities, all the results indicate that we have successfully explored a more suitable similarity structure for unsupervised cross-modal hashing from a distribution perspective. By combining collaborative consistency learning, \method{} effectively improves the image-text matching quality.
\end{enumerate}




